\documentclass[conference]{IEEEtran}
\IEEEoverridecommandlockouts
\usepackage[utf8]{inputenc} 
\usepackage[T1]{fontenc} 
\usepackage{cite}
\usepackage{amsmath,amssymb,amsfonts}
\usepackage{algorithmic}
\usepackage{graphicx}
\usepackage[small]{caption}
\usepackage{textcomp}
\usepackage{xcolor}
\usepackage{balance}
\usepackage{bbding}
\usepackage{pifont}
\usepackage{wasysym}
\usepackage{interval}
\usepackage[draft]{hyperref}
\usepackage{multirow}
\usepackage{paralist}

\usepackage{placeins} 
\def\BibTeX{{\rm B\kern-.05em{\sc i\kern-.025em b}\kern-.08em
    T\kern-.1667em\lower.7ex\hbox{E}\kern-.125emX}}

\newcommand{\cmark}{\ding{51}}%

\newcommand{\norm}[1]{\left\lVert#1\right\rVert^{2}}

\usepackage{siunitx}

\def\tablename{Table}

\begin{document}

\title{Pedestrian Tracking by Probabilistic Data Association and Correspondence Embeddings}

\author{
\IEEEauthorblockN{Borna Bićanić \qquad
				  Marin Oršić \qquad
				  Ivan Marković \qquad
				  Siniša Šegvić \qquad
				  Ivan Petrović
			  	 }
\IEEEauthorblockA{
University of Zagreb, Faculty of Electrical Engineering and Computing\\
Email: {\{borna.bicanic, marin.orsic, ivan.markovic, sinisa.segvic, ivan.petrovic\}@fer.hr}
}
}

\maketitle

\begin{abstract}
This paper studies the interplay between kinematics (position and velocity) and appearance cues for establishing correspondences in multi-target pedestrian tracking.
We investigate tracking-by-detection approaches based on a deep learning detector,
joint integrated probabilistic data association (JIPDA),
and appearance-based tracking of deep correspondence embeddings.
We first addressed the fixed-camera setup by fine-tuning a convolutional
detector for accurate pedestrian detection and combining it with kinematic-only JIPDA.
The resulting submission ranked first on the 3DMOT2015 benchmark.
However, in sequences with a moving camera and unknown ego-motion,
we achieved the best results by replacing kinematic cues with
global nearest neighbor tracking of deep correspondence embeddings.
We trained the embeddings by fine-tuning features
from the second block of ResNet-18 using angular loss extended by a margin term.
We note that integrating deep correspondence embeddings directly in JIPDA did not bring significant improvement.
It appears that geometry of deep correspondence embeddings
for soft data association needs further investigation in order to obtain the best from both worlds.
\end{abstract}

%

\section{Introduction}
\label{sec:intro}

The application of multi-target tracking (MTT) algorithms is found today in many different areas, such as autonomous vehicles, robotics, air-traffic control, video surveillance and many other.
Tracking and state estimation of multiple moving objects gives rise to many challenges compared
to classical estimation, like time-varying number of targets, false alarms, occlusions, and missed detections.
Additional level of complexity is brought by unknown association between measurements and targets.
In visual MTT this can be alleviated by leveraging appearance of the detected targets in images.
Thus a visual tracking approach usually combines the following three steps:
\begin{inparaenum}[(i)]
	\item detection of objects in images, in our case pedestrians,
	\item computation of appearance based metrics for association, and
	\item a tracking algorithm consolidating the previous two steps into the MTT framework.
\end{inparaenum}
\begin{figure}[!t]
\centering
\includegraphics[width=0.45\columnwidth,trim={530px 380px 80px 90px},clip]{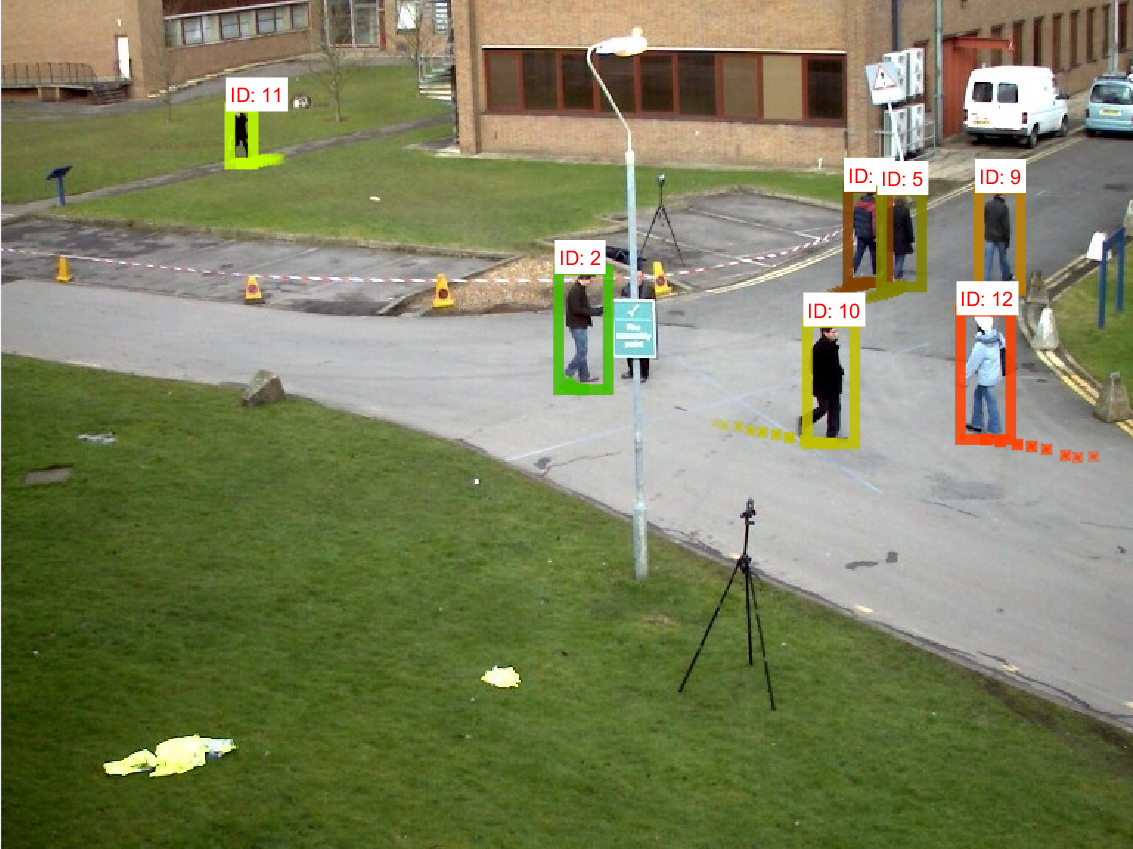}
\includegraphics[width=0.45\columnwidth,trim={530px 380px 80px 90px},clip]{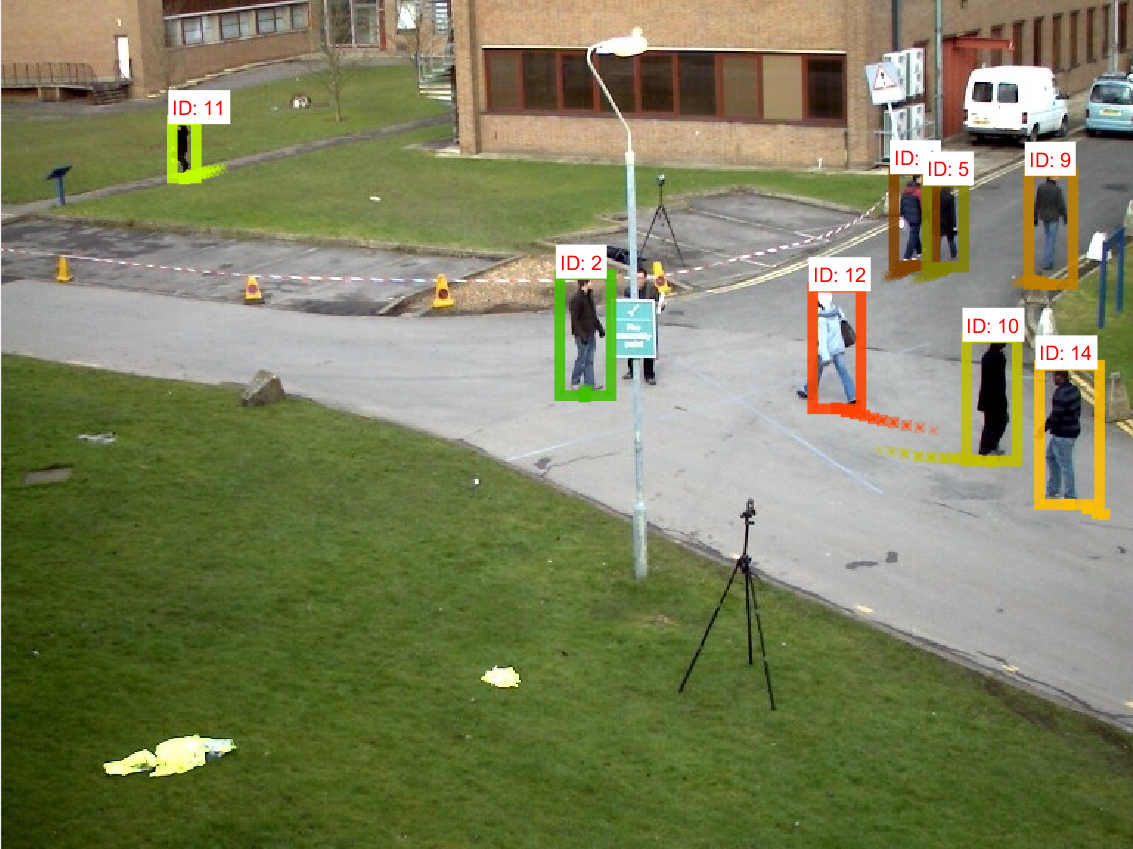}
\\[0.12cm]
\includegraphics[width=0.45\columnwidth,trim={300px 180px 360px 0px},clip]{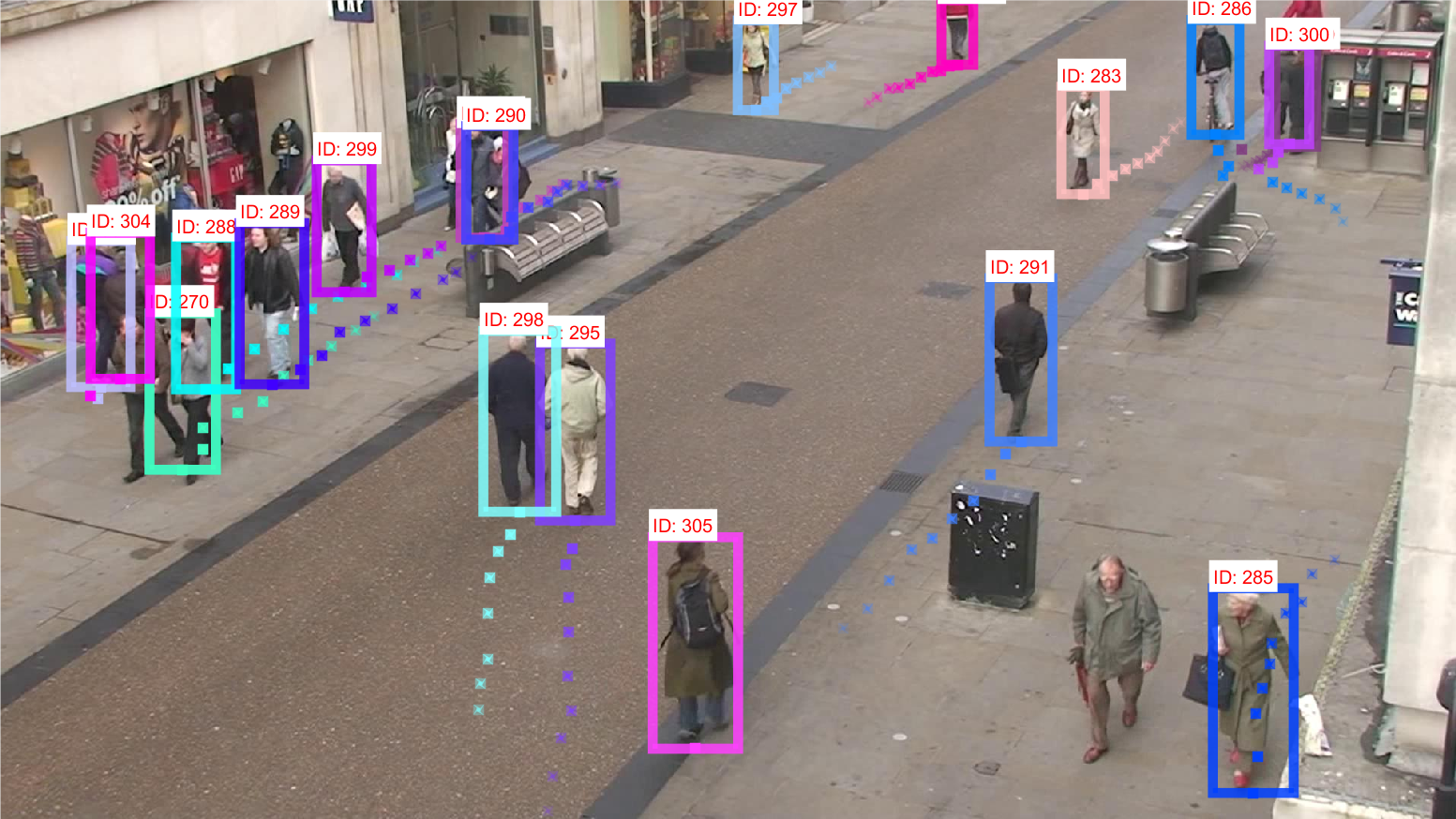}
\includegraphics[width=0.45\columnwidth,trim={300px 180px 360px 0px},clip]{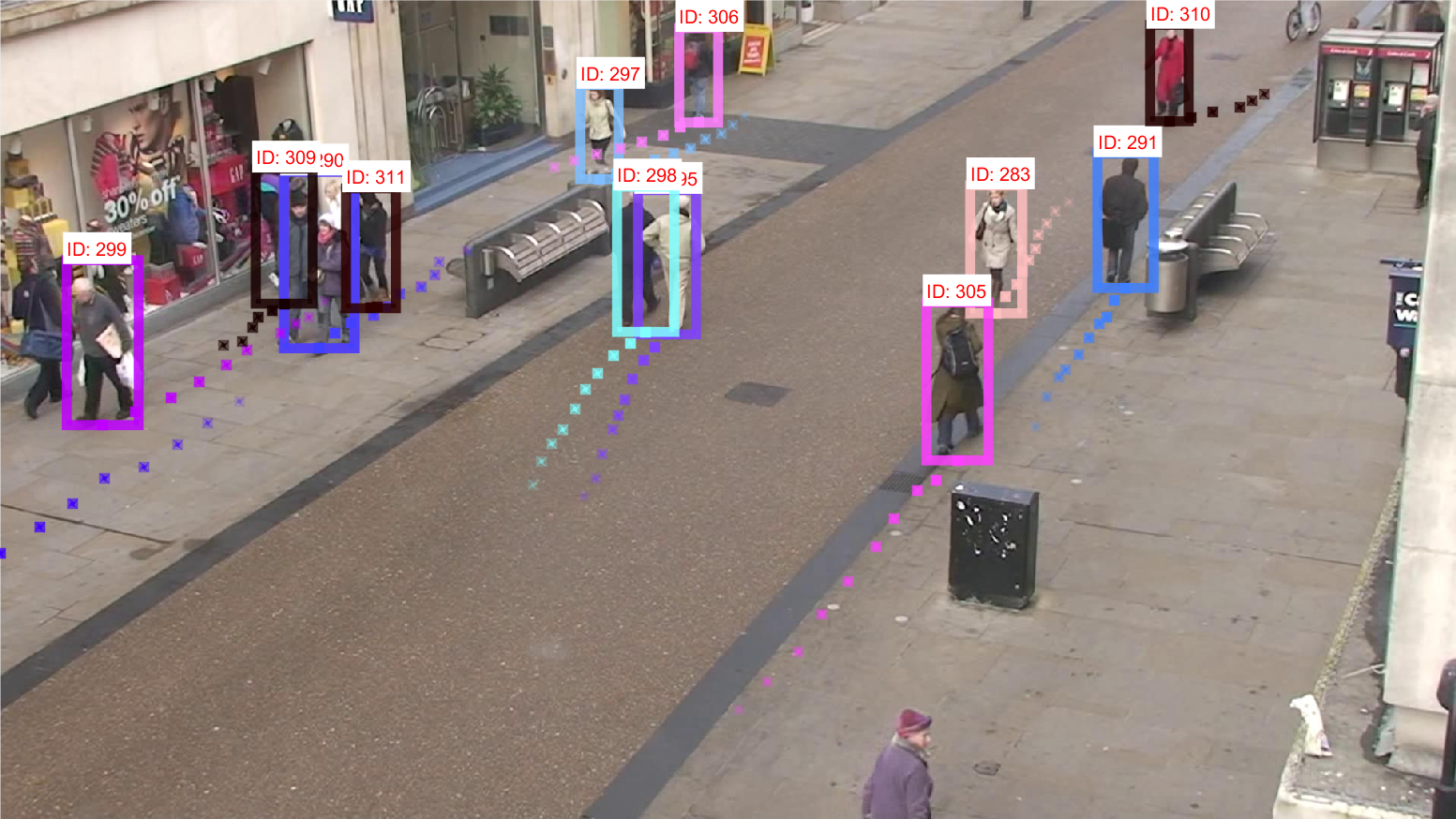}
\caption{Pedestrian tracking on 3DMOT2015 sequences, PETS09-S2L1 (up) frame 125 (left) and frame 138 (right),
AVG-TownCentre (down) frame 341 (left) and frame 351 (right).  \label{fig:mot15_3d}}
\vspace{-0.5cm}
\end{figure}

Pedestrian MTT methods require detections as inputs and deep convolutional  models are particularly suitable for the task.
Pre-training deep models on large datasets is shown to have a great regularization effect. 
Datasets like ImageNet \cite{russakovsky15ijcv} and COCO \cite{lin14corr} hold great generalization potential which is available through pre-training on such large collections of annotated data.
However, fine-tuning of a multi class object detector to detect pedestrians is not a straightforward task.
Limitations in model vertical receptive field, noise in bounding box annotations, and annotation errors are common issues \cite{zhang16cvpr}.
Fine-tuning on homogeneous video sequences incurs high overfitting risk,
thus diversity in training data should be targeted to improve generalization.
Given that, fine-tuning a multiclass object detector for the task of pedestrian
detection is possible by training on a dataset like CityPersons \cite{zhang17cvpr}.
This dataset contains diversity on multiple axes,
such as person identity, clothing, pose, occlusion level etc.

Correspondence embeddings can be useful for a target association problem such as pedestrian tracking,
since they are trained to measure similarity between images.
The pioneer of approaches for deep metric learning used siamese networks \cite{bromley94nips},
while triplet networks are considered as an improvement \cite{hoffer15simbad}.
This is due to a slight, but impactful modification of the loss,
which ensures better alignment between the similarity in the embedded space and the likeliness of correspondence.
A body of work analyses and improves triplet loss functions.
Convergence issues of metric learning using triplet loss are alleviated using N-pair loss,
which compares a positive example to N-1 negatives \cite{sohn16nips}.
Rather than focusing on pairwise distances in metric space,
angular loss \cite{wang17iccv} minimizes the angle at the negative point of the triplet.
This has a positive effect on quality of learning,
since angles are insensitive to changes in scale.
Our approach follows previous work which utilizes
segmentation masks to make the appearance embedding
less sensitive to occlusions and changes in the background \cite{song18cvpr}.

A good overview of the current state-of-the-art of the MTT algorithms can be found in \cite{Vo2015}, where authors
consider three different approaches to the MTT problem:
\begin{inparaenum}[(i)]
\item probabilistic data association,
\item multiple hypothesis tracking, and
\item random finite set approach.
\end{inparaenum}
In probabilistic data association (PDA) \cite{Bar-Shalom1975,Fortmann1983,Musicki2002} tracking methods, the measurement association uncertainty is untangled by soft assignment.
The first such methods for single and multiple target tracking were the PDA filter \cite{Bar-Shalom1975} and the joint probabilistic data association (JPDA) filter  \cite{Fortmann1983}, respectively.
However, both approaches assume known and constant target number and need some heuristics for track initialisation and termination.
The integrated PDA (IPDA) \cite{Musicki1994} and joint integrated PDA (JIPDA) \cite{Musicki2002} alleviate this issue by estimating the targets existence probability together with its states, thus providing a natural method for automatic track initialisation and termination.
Unlike PDA methods, multiple hypothesis tracking (MHT) algorithms \cite{Reid1979,Blackman2004} generate hypotheses for different associations and the decision about which of the hypotheses is correct is postponed until new data is collected.
Somewhat more recent approaches are based on the random finite set (RFS) paradigm \cite{Goodman1997}.
Based on the RFS theory, the closed-form first moment approximation of the RFS filter, the Gaussian mixture probability hypothesis density (GM-PHD) filter was presented in \cite{Vo2006a}, and since then other novel approaches have been proposed \cite{Mahler2007a,Reuter2014,Krishanth2017}.

Visual pedestrian tracking can be implemented by consolidating an object detector,
appearance based association metric, and a suitable MTT approach.
In \cite{Wang2018} authors track multiple pedestrians using a Rao-Blackwellized particle filter with track management based on detection association likelihoods.
Therein, the authors augment the state vector of tracked objects by an appearance based deep person re-identification vector \cite{Wang2018a} and compute data association probability by multiplying conditionally independent position and appearance association likelihoods.
The authors report that adding appearance information reduced the number of identity switches and increased slightly the overall tracking score; however, tracking using just the appearance, without position information, showed to perform quite poorly.
In \cite{klinger15isprs} probabilistic models were incorporated into a track-by-detection approach using prior knowledge of a static scene, describing pedestrian state using position, height and width in
world coordinates. Such approach lacks information on pedestrian appearance to correctly handle interactions between pedestrians in crowded scenes.
The MOANA approach \cite{tang19ieee} uses hand-crafted features to represent appearance and resolve
situations when a candidate observation is spatially close to other objects.

In this paper we present a pedestrian tracking-by-detection approach based on a deep learning detector combined with the JIPDA and an appearance-based tracker using deep correspondence embeddings.
A convolutional neural network detector was pretrained on the COCO dataset for accurate pedestrian detection and serves as the input for the JIPDA based tracking algorithm where the state consists only of pedestrian kinematic cues (positions and velocities).
The proposed pedestrian tracker with kinematic cues currently ranks first on the 3DMOT2015 online benchmark \cite{Leal-Taixe2015} that contains sequences with a static camera (cf. \figurename~\ref{fig:mot15_3d}).
In order to enable pedestrian tracking in sequences containing camera motion, under the assumption that camera ego-motion is not available, kinematic parameters in 3D need to be exchanged for appearance cues based on a deep correspondence metric in the image space.
We therefore learn a correspondence embedding and leverage
it for association across video frames
using the global nearest neighbor approach (GNN).
In the end, we compare GNN tracking of correspondence embeddings
with the JIPDA tracker based on kinematic cues (position and velocity).

This paper is organized as follows.
In \autoref{sec:marin} we describe models we used
to detect pedestrians as well as calculate correspondence embeddings.
We outline our method for probabilistic association in \autoref{sec:jipda}.
Results on the online MOT benchmark as well as validation experiments are
presented in \autoref{sec:results}.
Finally, we give a summary of our accomplishments and findings in \autoref{sec:conclusion}.

\section{Detection and appearance representation}
\label{sec:marin}

We detected pedestrians with the Mask R-CNN algorithm trained on a
suitable blend of public datasets.
We cropped and scaled the bounding boxes, applied instance segmentation masks
and processed them with a separate model trained with a metric loss.
This resulted in correspondence embeddings which we used as
descriptors in appearance-only tracking.
We describe the details in the following subsections.

\subsection{Pedestrian detection}
Mask R-CNN \cite{he17iccv} is an extension of the
Faster R-CNN \cite{ren15nips} object detector.
It consists of two stages:
\begin{inparaenum}[(i)]
\item finding regions of interest
(RoIs) using region proposal network (RPN) and
\item classification of the proposed RoIs
and bounding box regression.
\end{inparaenum}
Mask R-CNN enhances the second stage by
predicting segmentation masks
of RoIs provided by RPN.
By utilizing the RoIAlign operation
and attaining better representations
through learning segmentation masks,
Mask R-CNN surpasses Faster R-CNN on the task
of object detection. We adapted the multi-class
Mask R-CNN for pedestrian detection.

We chose the most suitable transfer-learning strategy by
performing validation experiments with a Mask R-CNN
detector trained on different datasets.
Fine-tuning from COCO to CityPersons
turned out to be the most appropriate course of action as
shown in detail in \autoref{sec:results}.
We believe this can be explained as follows.
Firstly, CityPersons
includes annotations with fixed aspect ratio (\emph{BB-full})
which are suitable to train occlusion invariant bounding box regression.
Secondly, we noticed that COCO people
are much more diverse than MOT
pedestrians due to numerous other contexts such as riding, driving, sitting down etc.
Furthermore, CityPersons inherits ground truth pixel-level masks
from the Cityscapes dataset \cite{cordts15cvpr}.
Presence of ground truth pixel level masks
is suitable for fine-tuning Mask R-CNN's mask head.

We adapted the pre-trained multi class Mask R-CNN \cite{he17iccv}
for pedestrian detection in two steps.
Firstly, we adapted the Mask R-CNN classification,
bounding box regression and mask prediction heads
to have two possible outputs: background and pedestrian.
We sliced the weights of the last layer of the classification head
in order to leave only the logits for the \textit{background}
and \textit{pedestrian} classes which we initialized with
weights of the corresponding COCO classes.
Secondly, we fine-tuned the resulting
model with ground truth bounding boxes
and segmentation masks from CityPersons.

\subsection{Deep correspondence embedding}
We represented pedestrian appearance with a metric embedding provided by a deep correspondence model.
Appearance of each pedestrian is represented by high-dimensional
embeddings in metric space.
Selection of the correspondence model is not straightforward.
We started form ResNet-18 \cite{he16cvpr} classification
architecture which consists of four residual blocks, from RB1 to RB4.
Features from RB4
are suitable for discriminating between different classes.
However, we found that features from earlier blocks are more beneficial for
differentiating between different person identities.
Therefore, we calculated embeddings from features
in the last convolutional layer of RB2.
Validation experiments suggest that these features contain more information about person appearance than features in any other residual block.
Furthermore, the last two blocks hold around $\SI{70}{\percent}$ of total
ResNet parameters. By getting rid of them, we
decreased the susceptibility to overfitting. At the same time,
it is possible to initialize the first two blocks with pre-trained weights
and profit from regularization induced by ImageNet.
We demonstrate the effectiveness of this approach
in more detail in \autoref{sec:results}.

Furthermore, we investigated the possibility of using segmentation
masks provided by Mask R-CNN to generate descriptors which
are robust to changes in object background and occlusions.
We experimented with two approaches for incorporating the
segmentation mask $M_\mathrm{S}$ into the correspondence embedding.
The first approach applies $M_\mathrm{S}$ to the input image.
The second approach uses $M_\mathrm{S}$
to mask the convolutional features.
The two approaches are not the same
since the latter approach preserves
some background influence due to
receptive field of the convolutional features.
Despite this, applying the mask to ResNet
features performed better in experiments.
We conjecture that this is due to
low resolution of the Mask-RCNN mask resulting in poor accuracy
when upsampled to RoI resolution.
Note that a segmentation mask can be interpreted as a dense
probability map that the corresponding pixel is foreground.
Therefore, one can suppress the background
by elementwise multiplication with the segmentation mask.

As mentioned before, we adapted the ImageNet pre-trained architecture
by taking only the first two residual blocks.
The features of the last residual block were passed to a $1 \times 1$
convolutional layer and masked
using the output of Mask R-CNN's segmentation head.
Finally, the correspondence embedding was produced
by global average pooling.

The model was trained using angular loss \cite{wang17iccv}.
We extended the angular loss by adding the margin term.
For a given reference embedding $\mathbf{r}$, a corresponding
embedding of the same identity $\mathbf{p}$ and
a negative embedding $\mathbf{q}$,
we calculated the angular loss \eqref{eq:angular},
where $m$ is the margin hyperparameter and
$\mathbf{c} = \frac{\mathbf{r} + \mathbf{p}}{2}$:
\begin{equation}
	\label{eq:angular}
	L_{\mathrm{ang}} = \max(0, m + \norm{\mathbf{r} - \mathbf{p}} - 4 \tan(\alpha)^{2} \norm{\mathbf{q} - \mathbf{c}}).
\end{equation}
Gradients of the angular loss push the negative
example away from the center of $\mathbf{p}$ and $\mathbf{r}$
examples in the $\mathbf{q} - \mathbf{c}$ direction.
This also minimizes $\mathbf{r}^{T} \mathbf{q}$ and $\mathbf{p}^{T} \mathbf{q}$
($\mathbf{r}$, $\mathbf{p}$ and $\mathbf{q}$ are unit vectors).

\subsection{Details of training correspondence embedding}

We trained the correspondence model on MOT2016 \cite{Milan2016a}.
We refrain from training on 2D MOT 2015 since
it does not include precise ground
truth data regarding occlusion level.
During training, we removed all training samples with occlusion level
greater than $\SI{50}{\percent}$.
We incorporated the following method for generating positive and negative samples.
We generated positive examples by taking
random detections less than 5
frames away from the reference example frame.
We generated negative examples by taking random
identity from the same sequence.
We sampled random easy negatives as
bounding boxes which do not intersect any ground truth bounding boxes.
This made the correspondence model more robust to pedestrian detector's
false negative outputs.
We chose the following sequences to serve as validation data:
\texttt{MOT16-02}, \texttt{MOT16-04}, \texttt{MOT16-05}.
The validation data was used for early stopping and tuning of
hyperparameters.
The output embedding vectors had 64 dimensions.
We used the Adam \cite{kingma14corr} optimizer with fixed learning rate of $10^{-4}$.
Weight decay was set to $10^{-4}$ for all parameters and the model was trained for 10 epochs.
During training and testing, we did not use whole images.
Instead, we cropped the detection bounding boxes and resized them
to the fixed resolution $224 \times 96$.

\section{Joint Integrated Probabilistic Data Association}
\label{sec:jipda}

Probabilistic data association algorithms use the soft assignment method to update the states of each individual target with all available measurements.
However, this results in a large number of possible joint associations events that have to be considered.
This challenge can be further aggravated if the targets are not well separated and the calculation of the a posteriori association probabilities may become intractable for practical application.
However, the number of events can be significantly reduced by a validation process in which the association
hypotheses that are very unlikely are discarded.
Furthermore, an efficient approximation of the JPDA was proposed in \cite{Hamid} with $m$ best joint associations; nevertheless, in this paper we consider the exact JIPDA algorithm since the targets are well separated and the
clutter rate is low.

JIPDA predicts the target state individually for each track.
If we construct an appropriate target motion model, then the state of each target can be propagated using the standard Kalman filter prediction step.
Additionally, the target existence probability prediction is given by
\begin{equation}
P_{k|k-1}(\mathcal{H}^j | Z^{1:k-1}) = P_S \, P_{k-1|k-1}(\mathcal{H}^j | Z^{1:k-1}) \text{,}
\end{equation}
where $P_S$ is the target survival probability, $Z^k = \lbrace z_i^k \rbrace_{i=1}^{m_k}$ is the set of all observations at time $k$ where $m_k$ is the number of observations at time $k$. $Z^{1:k}$ is the set of all observations up to and including time $k$ and $\mathcal{H}^j$ is the hypothesis that track $j$ exists.

Let $\nu_{i,j}=z_i-\hat{z}_j$ denote the innovation of the $i$-th measurement to the track $j$, where $z_i$ is the measurement and $\hat{z}_j$ is the predicted measurement for track $j$.
Time superscripts are omitted here for clarity.
The target state is then corrected by the Kalman filter update equation
\begin{equation}
\hat{x}_{k|k} = \hat{x}_{k|k-1} + K_{j,k} \nu_j \text{,}\label{eq:jipda_state_update}
\end{equation}
where $\nu_j = \sum_{i = 1}^{m_k} \beta_{i,j} \, \nu_{i,j}$ is the weighted innovation, $\beta_{i,j}$ are posterior association probabilities, and $K_{j,k}$ is the Kalman gain for target $j$.
The update of the covariance matrix slightly differs from the original Kalman update step \cite{Fortmann1983}
\begin{align}
P_{j,k|k} & = P_{j,k|k-1} - (1 - \beta_{0,j})K_{j,k} S_{j,k} K_{j,k}^T + P_{j,k} \text{,} \label{eq:jipda_cov_update}\\
P_{j,k} & = K_{j,k} \left[{ \sum_{i = 1}^{m_k} \beta_{i,j} \nu_{i,j} \nu_{i,j}^T - \nu_j \nu_j^T }\right] K_{j,k}^T
 \text{,}
\end{align}
where $S_{j,k}$ is the innovation covariance of the target $j$.

The combinatorial computational complexity of the JIPDA can be alleviated by discarding the assignment hypotheses that are unlikely.
Since the innovation of the measurement is a zero-mean normal distribution, the measurement validation can be achieved by selecting only those measurement that lie in the confidence ellipsoid of the target \cite{Bar-Shalom1975}.
%
A priori likelihood function of a measurement $i$ given state of a target $j$ after validating with the gating probability $P_G$ is given by
\begin{equation}
g_{i,j} \triangleq g(z_i \mid \hat{z}_j) = P_G^{-1} \, \mathcal{N}(z_i;\hat{z}_j, S_j) \text{,} \label{eq:likelihood}
\end{equation}
when $z_i$ is inside the validation gate and zero otherwise.
%

To calculate a posteriori association probabilities $\beta_{i,j}$, all possible joint association events must be considered.
In each event, one target can be associated with at most one detection, and each detection cannot be assigned to more than one target.
Let $\mathcal{A}$ denote the set of all joint association events.
Since those events are exhaustive and mutually exclusive, the probability of the joint event $\mathcal{A}_i$ is given by \cite{Musicki2002}
\begin{align}
P(\mathcal{A}_i \mid Z^{1:k}) = C  & \times \prod_{j \in T_0} \left(1 - P_D \, P_G \, P_{\mathcal{H}^j}\right) \nonumber \\
& \times \prod_{j \in T_1} \left( P_D \,P_G \, P_{\mathcal{H}^j} \, g_{i,j} \, \lambda^{-1} \right) \label{eq:joint_event_prob} \text{,}
\end{align}
where $C$ is the normalization constant, $P_D$ is detection probability, $\lambda$ is clutter density, $T_0$ and
$T_1$ are sets of tracks assigned with no measurements and with one measurement in $\mathcal{A}_i$ and $P_{\mathcal{H}^j} = P_{k|k-1}(\mathcal{H}^j | Z^{1:k-1})$.

Let $\mathcal{H}_i^j$ be the hypothesis that the measurement $i$ belongs to target $j$ and
$\mathcal{H}_\emptyset^j$ the hypothesis that the target $j$ was not detected.
The a posteriori probabilities of individual track existence and measurement association can be obtained by \cite{Musicki2002}
\begin{equation}
P(\mathcal{H}^j,\mathcal{H}_i^j \mid Z^{1:k}) = \!\!\!\! \sum_{\mathcal{A} \in \mathcal{A}(i,j)} \!\!\!\! P(\mathcal{A} \mid Z^{1:k}) \text{,} \label{eq:p_exist_detected_i}
\end{equation}
\vspace{-0.25in}
\begin{align}
P(\mathcal{H}^j, \mathcal{H}_\emptyset^j \mid Z^{1:k}) = & \frac{(1 - P_D P_G)P_{\mathcal{H}^j} }{1 - P_D P_G P_{\mathcal{H}^j}}  \nonumber \\[0.1cm]
 & \times \!\!\!\!\!\! \sum_{\mathcal{A} \in \mathcal{A}(\emptyset,j)} \!\!\!\! P(\mathcal{A} \mid Z^{1:k}) \text{,} \label{eq:p_exist_not_detected}
\end{align}
where $\mathcal{A}(i,j)$ is the set of all events that assign measurement $i$ to track $j$, while $\mathcal{A}(\emptyset,j)$ is the set of all events in which track $j$ was missed.
Given probabilities \eqref{eq:p_exist_detected_i} and \eqref{eq:p_exist_not_detected}, the a posteriori track existence probability is computed as
\begin{align}
P_{k|k}(\mathcal{H}^j \mid Z^{1:k}) = & P(\mathcal{H}^j, \mathcal{H}_\emptyset^j \mid Z^{1:k}) \nonumber \\[0.1cm]
& + \!\!\!\!\!\!\!\! \sum_{i \in \lbrace \mathcal{M}_{i,j} = 1 \rbrace } \!\!\!\!\!\!\!\! P(\mathcal{H}^j,\mathcal{H}_i^j \mid Z^{1:k}) \text{,} \label{eq:a_posteriori_existence_prob}
\end{align}
where $\mathcal{M}_{i,j}$ is the element of the validation matrix, while a posteriori association probabilities are given by
\begin{equation}
\beta_{i,j} = \frac{P(\mathcal{H}^j, \mathcal{H}_i^j \mid Z^{1:k})}{P_{k|k}(\mathcal{H}^j \mid Z^{1:k})} \text{.} \label{eq:beta_a_posteriori}
\end{equation}

\section{Experimental Results}
\label{sec:results}

\subsection{Validating the detection and correspondence model}
\label{sec:results_ablation}

For validation experiments we studied the impact of training Mask R-CNN on different
combinations of training datasets and we carefully analyzed the design possibilities to find the most suitable correspondence embedding.
Here, we describe several validation studies and comment on the results.

\paragraph{Fine tuning Mask R-CNN}
After having little success in transfer learning from COCO to MOT in our preliminary experiments, we performed validation experiments by training just on CityPersons, just on COCO,
and on both datasets, achieving average precision of 45.1, 53.3, and 57.0, respectively.
Fine-tuning on CityPersons is suitable to distinguish
between pedestrians and other people.
Also, bounding boxes generated by Mask R-CNN
trained using \emph{BB\_full} annotations from CityPersons
are a better fit for detection of MOT pedestrians.
All our detection experiments feature
Mask-RCNN based on ResNet-50 FPN.

\label{par:Mask R-CNN}

\paragraph{Using segmentation maps}
The impact of using segmentation masks
is shown in \autoref{tab:ablation_mask}, where \emph{IDs} denote the
number of identity switches, while \emph{IDs\dag} shows evaluation on ground truth bounding boxes.
There are more \emph{IDs} when evaluating on ground truth because no fragmentations are present.
The models were trained on the MOT2016 train dataset, while evaluation was performed using an appearance based GNN
on 2DMOT2015 train.
We showed that segmentation masks
generated by Mask R-CNN benefit the correspondence
model by alleviating impacts of background and occlusions.
First, we trained a baseline correspondence model which
did not use segmentation masks.
Secondly, we trained two correspondence
models improved by segmentation masks.
The first model masks the input image.
The second model masks the final
feature map before the global average pooling operation.
We witnessed an improvement in tracking with the latter approach.

\begin{table}[t]
	\centering
	\caption{
		Applying segmentation masks at image vs feature level.
	}
	\label{tab:ablation_mask}
	\begin{tabular}{lccc}
		\hline\hline\\[-0.25cm]
		masked tensor       & IDs\dag      & IDs          &          MOTA \\
		\hline\\[-0.25cm]
		--                  & 507          & 404         &          53.6 \\
		input image         & 420          & 337         &          53.8 \\
		final conv features & \textbf{328} & \textbf{291} & \textbf{53.9} \\
		\hline\hline
	\end{tabular}
\end{table}

\paragraph{Residual blocks}

\begin{figure}[t]
	\centering
	\includegraphics[width=0.95\columnwidth]{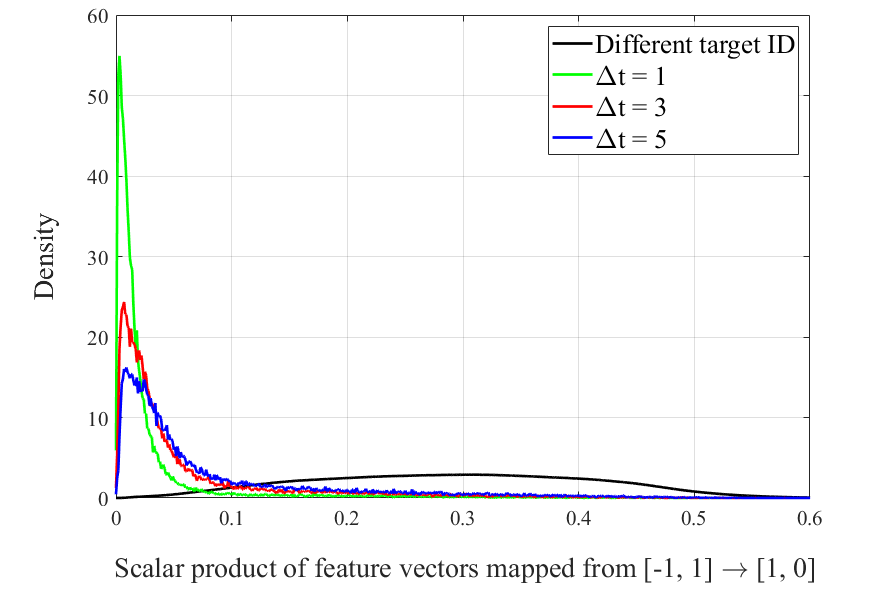}
	\caption{Distribution of scalar products of the deep embeddings mapped to interval $[0,1]$.
	Black line is the distribution of feature vector scalar products
	which do not belong to the same object.
	Red, green and blue lines show distributions
	of feature vector scalar products of the same object
	at consequent time steps.
	Feature vectors were evaluated on ground truth bounding boxes.}
	\label{fig:descriptor_similarity}
\end{figure}

Our final model uses only the first two residual blocks
of an ImageNet pre-trained ResNet-18. This design choice
is supported by experiments shown in \autoref{tab:ablation_resblocks}.
In each experiment, we used one additional residual block.
We trained the model on MOT2016 and evaluate tracking using a
position-agnostic GNN approach.
The results complement our initial hypothesis that for describing appearance,
abstract features like ones in the output of a full ResNet model may
not be beneficial.
In \figurename~\ref{fig:descriptor_similarity} we can see how the appearance similarity is distributed throughout the frames by looking at the similarity score of appearance vectors of the same object separated in time.
The results show that even for a separation of five time steps the similarity of most appearance vector is preserved, with clear separation from the other objects.

\begin{table}[b]
	\centering

	\caption{
		Validation of the model architecture.
		\emph{RB*} designates a resblock, while \emph{\#params} shows the total parameter count.
	}
	\label{tab:ablation_resblocks}
	\begin{tabular}{ccccccc}
		\hline\hline\\[-0.25cm]
		RB1    & RB2    & RB3    & RB4    &\#params& IDs\dag    & IDs         \\
		\hline\\[-0.25cm]
		\cmark &        &        &        & 161.7K &393         & 458         \\
		\cmark & \cmark &        &        & 691.3K &\textbf{328}& \textbf{291}\\
		\cmark & \cmark & \cmark &        & 2.8M   &416         & 398         \\
		\cmark & \cmark & \cmark & \cmark & 11.2M  &1271        & 687         \\
		\hline\hline
	\end{tabular}
\end{table}

\subsection{Pedestrian tracking evaluation}
\label{sec:results_mot3d}

To track the states of individual targets we used the constant velocity motion model with the state vector of the targets given by $x = [x_p,\, \dot{x}_p,\, y_p,\, \dot{y}_p]^T$.
We set the process and measurement noise deviations to $\sigma_q = \SI{0.836}{\meter\per\second\squared}$ and $\sigma_r = \SI{0.141}{\meter}$.
Target survival and detection probabilities were $P_S = 0.999$ and $P_D = 0.990$.
Measurement gating probability was $P_G = 0.990$.
False alarm rate was set to $\lambda = 15 V^{-1}$, where $V$ is the surveillance area. 
New targets were initialized for the measurements whose a posteriori association probability of not being associated to any of the existing targets was
\begin{equation*}
1 - \sum_j \beta_{i,j} > 0.7 \text{.}
\end{equation*}
Initial existence probability for new targets was set to $w_{\text{init}} = 0.65$
and the target was confirmed when its existence probability exceeded threshold $w_{\text{confirm}} = 0.85$.
Since nothing could be inferred about the new target's velocity from only one measurement,
it was assumed to be zero, but the initial covariance matrix of the target was inflated so that the state of the target converges to the actual value when the new measurements arrived.
Targets were terminated when their existence probability fell below the threshold $w_{\text{delete}} = 0.003$.
To improve tracking performance we discarded all detections with confidence score below the threshold $t = \SI{95}{\percent}$.

\begin{table}[!t]
	\centering
	\caption{
		\emph{MOTChallenge} benchmark results for \emph{3DMOT2015} category, proposed method is \emph{MCN\_JIPDA},
		$\uparrow$ denotes that higher is better and $\downarrow$ that lower is better,
		$^\ast$ denotes that the work is still unpublished. Best score for each metric is in boldface.}
	\label{tab:mot3d}
	\begin{tabular}{lccccc}
		\hline\hline\\[-0.25cm]
		Tracker                         &  MOTA$\uparrow$ &  MOTP$\uparrow$ &  FP$\downarrow$ &  FN$\downarrow$ &  IDs$\downarrow$ \\ 
		\hline\\[-0.25cm]
		MCN\_JIPDA                      &  \textbf{55.9}  &  \textbf{64.0}  & 2,910           & \textbf{4,011}  & 486              \\ 
		MOANA \cite{tang19ieee}         &  52.7           &  56.3           & 2,226           & 5,551           & \textbf{167}     \\ 
		DBN \cite{klinger15isprs}       &  51.1           &  61.0           & 2,077           & 5,746           & 380	           \\ 
		GPDBN\cite{klinger17isprs}      &  49.8           &  62.2           & \textbf{1,813}  & 6,300           & 311	           \\ 
		GustavHX$^\ast$                 &  42.5           &  56.2           & 2,735           & 6,623           & 302	           \\ 
		\hline\hline
	\end{tabular}
\end{table}

The tracking results are shown in \autoref{tab:mot3d}, where we can see that the proposed kinematic cues based JIPDA with the Mask R-CNN detector ranked first on the 3DMOT2015 dataset that contains static camera sequences.
The table shows results for the test sequence, while on the train sequences the tracker obtained MOTA 80.6 and MOTP 69.1.
Our method did produce a higher number of identity switches compared to MOANA, since we did not use appearance cues and our detector has higher recall than public detections.
The tracking performance could be further improved by using interacting multiple model \cite{DeFeo1997}
instead of a constant velocity Kalman filter and by taking unresolved measurements into account as proposed in  \cite{Svensson2012}.

In \autoref{tab:results_mot2d_sequences}, which compares the kinematic cues based JIPDA with deep detections to the deep correspondence metric based GNN,
we can see that both trackers show roughly the same performance
for static camera sequences and tracking in the image space,
while the kinematic based JIPDA is not appropriate for moving camera with unknown motion.
Augmenting the state space with deep correspondence embeddings directly within a soft data association approach such as JIPDA did not result in increased tracking accuracy in our experiments.
It remains an interesting venue of future work to investigate the correspondence embeddings space geometry and utilize the findings in soft data association approaches.

\begin{table}[!t]
	\centering
	\caption{Comparison of a kinematic based JIPDA and appearance based GNN on 2DMOT2015 train sequences.}
	\label{tab:results_mot2d_sequences}
	\begin{tabular}{clccc}
	\hline\hline\\[-0.25cm]
	Cam                                               & Sequence           & JIPDA                & Appearance GNN       \\
	\hline\\[-0.25cm]
	\multirow{6}{*}{\rotatebox[origin=c]{90}{Static}} &ADL-Rundle-6        & 58.4                  & 58.4                 \\
	                                                  &KITTI-17            & 58.3                  & 56,1                 \\
	                                                  &PETS09-S2L1         & 79.8                  & 78.8                 \\
	                                                  &TUD-Campus          & 78.3                  & 79.4                 \\
	                                                  &TUD-Stadtmitte      & 81.0                  & 81.6                 \\
	                                                  &Venice-2            & 46.0                  & 47.1                 \\
	\hline\\[-0.25cm]
	\multirow{5}{*}{\rotatebox[origin=c]{90}{Moving}} &ADL-Rundle-8        & --                    & 49.5                 \\
	                                                  &ETH-Bahnhof         & --                    & 29.4                 \\
	                                                  &ETH-Pedcross2       & --                    & 58.0                 \\
	                                                  &ETH-Sunnyday        & --                    & 62.8                 \\
	                                                  &KITTI-13            & --                    & 40.8                 \\[0.05cm]
	\hline\\[-0.2cm]
	\multicolumn{2}{c}{Total}                                              & --                    & 53.8                 \\[0.05cm]
	\hline\hline
	\end{tabular}
\end{table}

\section{Conclusion}
\label{sec:conclusion}
In this work we have proposed an online pedestrian tracking
method based on JIPDA and deep models for pedestrian detection
and correspondence embedding.
We have demonstrated how a COCO pre-trained Mask R-CNN
can be adapted for accurate pedestrian detection.
Furthermore, we incorporated segmentation masks to
improve the correspondence model embeddings.
Our correspondence embedding uses
masked features from the second residual block of ResNet-18
in order to focus on low-level foreground appearance
and reduce the parameter count.
The features are pre-trained on ImageNet
and fine-tuned with the angular loss.
We achieve our best results on the 3DMOT2015 benchmark
by combining Mask R-CNN detection and JIPDA.
Our submission achieves MOTA 55.9
and ranks \#1 at the time of writing this manuscript.
Suitable directions for future work include
integrating correspondence embeddings within JIPDA and
investigating the geometry of such soft data association.

\section*{Acknowledgement}
This work has been supported in part by the European Regional
Development Fund under the project ”System for increased
driving safety in public urban rail traffic (SafeTRAM)”
under Grant KK.01.2.1.01.0022
and in part by the Ministry of Science and Education of the Republic of Croatia
under the project Rethinking Robotics for the Robot Companion of the future (RoboCom++).
The Titan X used in experiments was donated by NVIDIA Corporation.

\balance
\bibliographystyle{IEEEtran}
\bibliography{main}

\end{document}